\documentclass[10pt,twocolumn,letterpaper]{article}

\usepackage{cvpr}
\usepackage{times}
\usepackage{epsfig}
\usepackage{graphicx}
\usepackage{amsmath}
\usepackage{amssymb}
\usepackage{multirow}

\usepackage[breaklinks=true,bookmarks=false]{hyperref}

\cvprfinalcopy 

\def\cvprPaperID{****} 

\begin{document}

\title{Sketch-based Normal Map Generation with Geometric Sampling}

\author{Yi He\\
JAIST\\
Ishikawa, JAPAN\\
{\tt\small s2010035@jaist.ac.jp}
\and
Haoran Xie\footnote{}\\
JAIST\\
Ishikawa, JAPAN\\
{\tt\small xie@jaist.ac.jp}
\and
Chao Zhang\\
University of Fukui\\
Fukui, JAPAN\\
{\tt\small zhang@u-fukui.ac.jp}
\and
Xi Yang\\
The University of Tokyo\\
Tokyo, JAPAN\\
{\tt\small earthyangxi@gmail.com}
\and
Kazunori Miyata\\
JAIST\\
Ishikawa, JAPAN\\
{\tt\small miyata@jaist.ac.jp}
}

\maketitle

\begin{abstract}
   Normal map is an important and efficient way to represent complex 3D models. A designer may benefit from the auto-generation of high quality and accurate normal maps from freehand sketches in 3D content creation. This paper proposes a deep generative model for generating normal maps from users’ sketch with geometric sampling. Our generative model is based on Conditional Generative Adversarial Network with the curvature-sensitive points sampling of conditional masks. This sampling process can help eliminate the ambiguity of generation results as network input. In addition, we adopted a U-Net structure discriminator to help the generator be better trained. It is verified that the proposed framework can generate more accurate normal maps.
\end{abstract}

\section{INTRODUCTION}
\label{sec:intro}  

Normal maps play an important role in Computer Graphics and Computer Vision research fields, but it is challenging for common users to design surface normal maps and infer accurately, especially in the early stage of a design process. The common users are used to drawing sketches to convey the idea for immediacy, and we believe that the automatic generation of normal maps from sketches can increase working efficiency in design processes. Current research progresses on content creation have been intensively explored with conditional Generative Adversarial Networks (GANs), such as vector field generation \cite{hu2019sketch2vf} and the normal map generation from sketches \cite{su2018interactive}. Inspired by these previous works, we aim to generate normal maps from sketch images with deep neural networks.

In this work, we proposed a novel network structure of normal map generation using conditional generative network structures.  The proposed network samples point hints by curvature distribution as a part of input data. Our method can eliminate the ambiguity in generated images and help to reduce generation errors. We demonstrate the effectiveness of our proposed method with both quantitative and qualitative experiments. The generated results show the capability in various kinds of tasks in the generation of low-error normal maps comparing with alternative approaches with conventional network structures including normal map generation with Sketch2Normal with random sampling \cite{su2018interactive} and Pix2Pix with conditional GANs \cite{isola2017image}.

\section{RELATED WORK}

\subsection{Sketch-Based 3D Shape Reconstruction}

Sketch-based modeling has been explored largely in computer graphics community, and many researchers have contributed to this topic \cite{olsen2009sketch}. This research topic is still challenging and popular in both computer graphics and computer vision fields. There are still numerous issues in sketch-based 3D reconstruction, such as ill-posed problems in model reconstruction from sketches. For 3D reconstruction, normal map plays an important role in 3D modeling applications. The generation of normal maps and depth maps from user sketches has been proposed frequently nowadays. 

Lun et al. proposed a deep learning based method to generate depth maps and normal maps from multi-view sketches, and then 3D models \cite{lun20173d}. Deepsketch2Face was proposed to generate detailed face models based on 2D sketch images \cite{han2017deepsketch2face}. Sketch2Normal aimed to generate normal maps that correctly represent 3D shapes through user interaction \cite{su2018interactive}. Inspired by this work, we aim to generate normal maps that can represent 3D shapes more accurately. In this work, the proposed method is more sensitive to the user sketch and the given geometric information to eliminate possible ambiguity.

\subsection{Image Translation}

For image style transfer and image translation issues, deep neural networks can achieve better results than conventional algorithms, which mainly uses handcrafted separate local image representation. Deep learning based approaches adopt a training dataset to provide input and output pairs to train the image translation functions. GANs networks are usually utilized to train and obtain the parameters of such objective functions.

The conventional conditional GANs can allow the generative models to generate demanded images \cite{mirza2014conditional}, which is greatly helpful for image translation and other related tasks. For the surface normal representation problem, Eigen and Fergus \cite{eigen2015predicting} and Wang et al. \cite{wang2015designing} contribute to the issue of coarse prediction of surface normals. Sketch2Normal can generate fine-grained normal maps of 3D object models \cite{su2018interactive}, they use deep neural networks to convert sketch to normal maps and adjust the generated normal maps by random user input to eliminate ambiguity. This approach utilizes pix2pix \cite{isola2017image}, one of the most versatile frameworks for image-to-image translation. To improve the previous works on normal map generation, we sample the surface curvature of the 3D model for training to eliminate ambiguity and generate an accurate normal map for 3D shape representation.

\section{PROPOSED METHOD}
\label{sec:title}
 \begin{figure*} [t]
   \begin{center}
   \begin{tabular}{c} 
   \includegraphics[height=8.5cm]{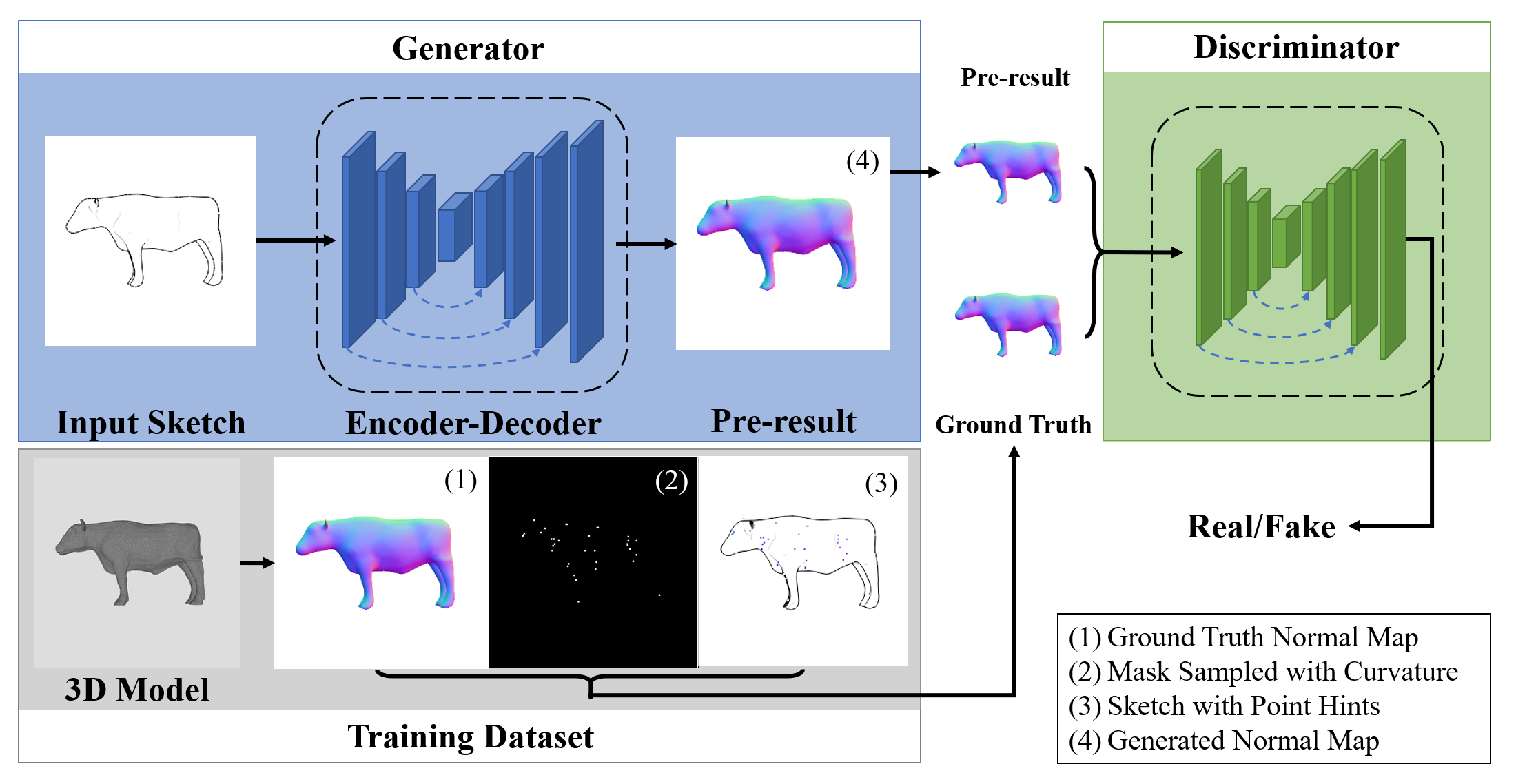}
   \end{tabular}
   \end{center}
   \caption[framework] 
   { \label{fig:framework} 
    The network structure of the proposed method. We first calculate the surface curvatures from ground truth and select the point hints correspondingly. The generator network input is the concatenation of the binary mask (2) and sketch contour (3) based on ground truth of the normal map (1). We use the intermediate result generated by the generator as input to a U-Net structure discriminator along with ground truth.}
   \end{figure*} 
   
In this work, we consider the sketch to normal map generation task as an image translation issue. We propose a novel normal map generation network as shown in Figure~\ref{fig:framework}. The proposed network is basically based on conditional GAN framework. First, we prepare the paired data of sketch and ground truth normal map in the training process. To incorporate the curvature sensitive point hints, we apply two additional losses to further regulate the image generation. We first calculate the surface curvatures from normal maps and sample the point hints by threshold as the binary mask. In the binary mask, the point value is set to be 1 in our implementation. In each training iteration, we replace the mask points of ground truth image on the generated results to ensure that the neighboring area can represent the specified normal information.
   
\subsection{Objective Function}

We use the conditional GAN network to map sketch image $x$ and a noise $z$ to a normal map of $y$ as equation (\ref{eq:map}). According to the conventional conditional GAN framework \cite{isola2017image}, our objective function is defined as equation (\ref{eq:cGAN}):
\begin{equation}
\label{eq:map}
G:x,z\rightarrow y \, .
\end{equation}
\begin{multline}
\label{eq:cGAN}
\min _{G} \max _{D} V(D, G)=E_{x \sim p_{sketch}, y \sim p_{nmap}}[\log D(y \mid x) \\
+E_{x \sim p_{sketch}, z \sim p_{rnoise}}[\log (1-D(G(z \mid x)))] \, .
\end{multline}

Here $x$ represents the input sketch image, and $y$ is the normal map associated with the sketch. $z$ denotes the random noise. In addition, $p_{sketch}$, $p_{nmap}$, and $p_{rnoise}$ represent the distributions of sketch, normal map, and random noise domains, respectively.

The generator function in the original GAN suffers from gradient vanishing\cite{goodfellow2014generative}, while wGAN can solve this problem by applying Wasserstein distances \cite{arjovsky2017wasserstein}. We mofify the objective function as follows:

\begin{multline}
L=E_{x \sim p_{sketch}, y \sim p_{nmap}}[D(y \mid x)]\\ -E_{\tilde{y} \sim p_{gnmap}}[D(\tilde{y} \mid x)] \, .
\end{multline}
\begin{equation}
\tilde{y}=E_{x \sim p_{sketch}, z \sim p_{rnoise}}[G(z \mid x)]
\end{equation}

Here, $\tilde{y}$ represents the generated normal map from the normal map domain $p_{gnmap}$ generated by input sketch images $x$.According to the previous work of context encoders \cite{pathak2016context}, the training process of the generator can be mixed by adding traditional loses, so we add $L_{L 1}$ here and the $L_{L 1}$ is defined as equation (\ref{eq:l1loss}). Our goal is to generate accurate normal map from the sketch image and eliminate ambiguity as much as possible. Therefore, we additionally provide $L_{\text {mask}}$ to assist in the training process of the network. We calculate the surface curvature from the normal map, and generate a mask with a threshold value to select points to assist the training process. We define the mask loss function $L_{\text {mask}}$ as equation (\ref{eq:maskloss}).

\begin{equation}
\label{eq:l1loss}
L_{L 1}=E_{y \sim p_{nmap}, \tilde{y} \sim p_{gnmap}}\left[\|y-\tilde{y}\|_{1}\right] \, .
\end{equation}
\begin{equation}
\label{eq:maskloss}
L_{\text {mask}}=\frac{1}{\sum \text { mask }} E_{y \sim p_{nmap}, \tilde{y} \sim p_{gnmap}}\left[\|y-\tilde{y}\|_{1} \odot \text { mask }\right]
\end{equation}

Therefore, the whole objective function is arranged as follows:

\begin{multline}
L=E_{x \sim p_{sketch}, y \sim p_{nmap}}[D(y \mid x)]\\ -E_{\tilde{y} \sim p_{gnmap}}[D(\tilde{y} \mid x)]\\ - \lambda_{L 1} L_{L 1}-\lambda_{\text {mask}} L_{\text {mask}} \, .
\end{multline}

\subsection{Network Design}
Figure~\ref{fig:framework} shows the network structure of the proposed normal map generation framework. For the network input, we concatenate the sketch image and mask image and generate a four-dimensional input of equal size to the original sketch. Then, the input is fed into the generator network (16 layers) for data training. For the generator network, we apply the Encoder-Decoder architecture \cite{hinton2006reducing}. Because it is desirable to pass low-level information across layers as well to guide the generation process for image translation \cite{isola2017image}, we apply the U-Net structure \cite{ronneberger2015u} in our generator. In this structure, we connect layer $i$ to layer $n-i$ after batch normalization, where $n$ refers to the number of network layers. Note that the U-Net structure for the discriminator can obtain a significant improvement in the generation results without modification of the generator \cite{schonfeld2020u}. Therefore, we built the discriminator with the same architecture and depth as the generator. For the interior of the network, each layer of network units is composed of a convolution layer, a batch normalization layer, and a leak ReLU unit. We apply RMSProp optimizer to train the network.


\section{Experiments}

For the preparation of the training dataset, we adopted the $FourLegs$ model published by Kalogerakis et al. \cite{kalogerakis2010learning}. We prepared the training dataset with paired sketch images and ground truth normal maps in 256$\times$256 pixels. We calculate the corresponding curvature distributions based on the normal maps, and every curvature map was processed with a threshold grayscale value of 127 and 126 at the same time to sample raw points and get 2 images including the raw point hints which represent where the most significant shape changes appeared. If we only use one threshold grayscale value to sample raw point hints, then the points we get will be a great amount and mostly sketch boundary points, which will not help in model training. These two images processed by threshold and including raw point hints are then subjected to an exclusive OR operation to generate a raw mask, and the curvature points selected by the raw mask are randomly dropped out with a probability of 0.95 to limit the number of curvature points. In our implementation, it required two days to train the model with NVIDIA GTX1070.

\begin{figure*}[!ht]
    \centering
    \includegraphics[height=8.2cm]{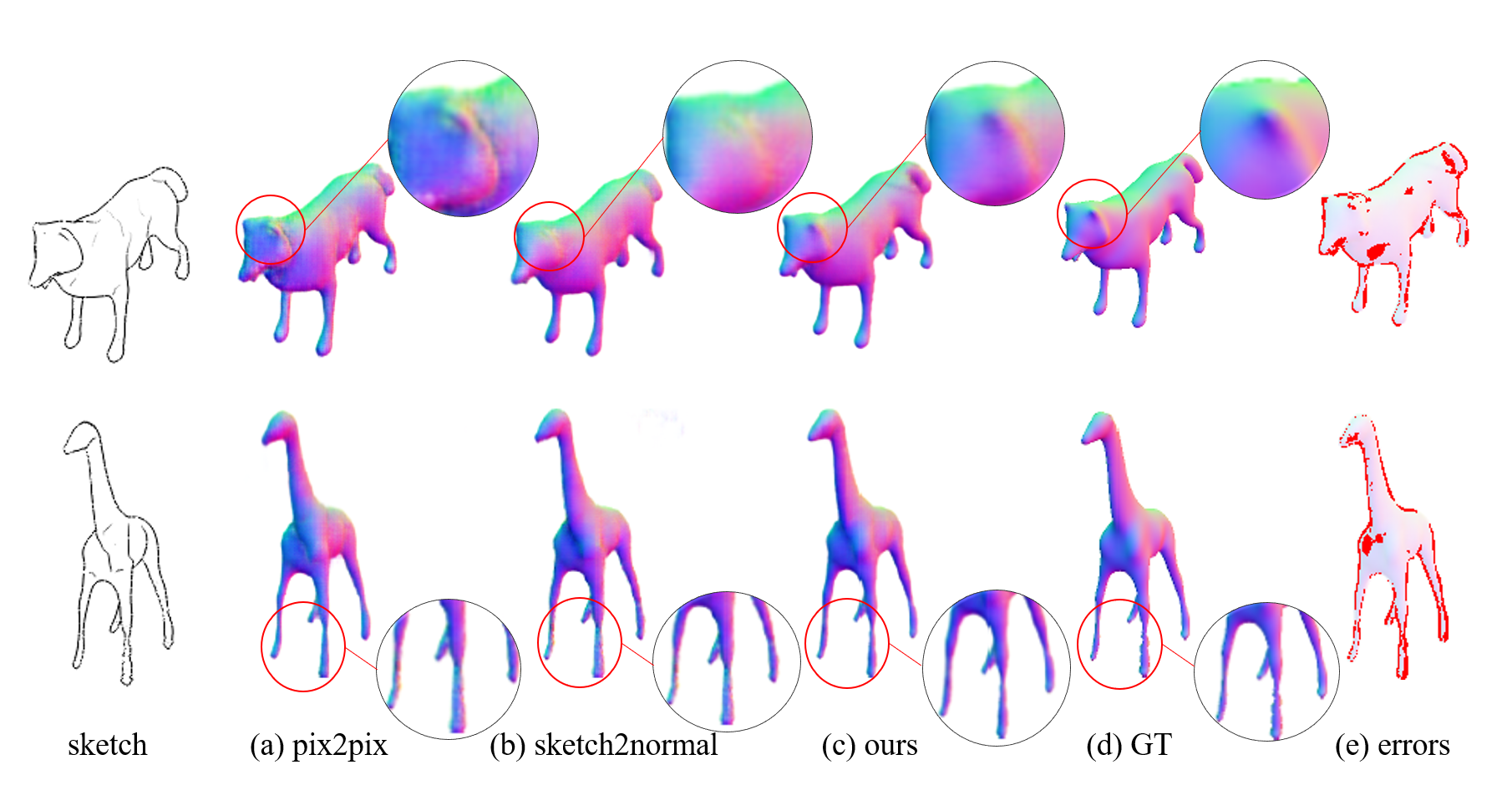}
    \caption{Comparison of  results generated by pix2pix(a), sketch2normal(b), our method(c), and ground truth(d). The sketch images used here are projections from the 3D models. We visualized the errors(e) between our results and ground truth.}
    \label{fig:results}
    \end{figure*}

We trained the model of Pix2Pix \cite{isola2017image} and Sketch2Normal \cite{su2018interactive} on our dataset. The evaluation of the proposed approach is conducted in both qualitative and quantitative experiments. We chose three metrics: L1, L2 distances, and angular difference to evaluate the proposed method with previous works as shown in Table~\ref{tab:evaluation}. We compared the results generated by Pix2Pix \cite{isola2017image} , Sketch2Normal \cite{su2018interactive} and ours to ground truth normal maps. Note that we focus on the normal regions on the averaged pixel-wise differences. The generated results by different methods are shown as Figure~\ref{fig:results}. Here the sketch images used here are projections from the 3D models. Figure 2(e) shows the errors between our results and ground truth. This indicates that the generated normal maps can express the user's design intentions with our proposed approach. The proposed approach can generate the most closest normal map to the normal distribution in ground truth.

    
    \begin{table}[ht]
    \caption{Metrics of normal map generation methods. } 
    \label{tab:evaluation}
    \begin{center}
    \begin{tabular}{r|ccc}
    \hline \multirow{2}*{Method} & &  Loss Type \\
    \cline{2-4} ~ & Angular & L1 & L2 \\
    \hline pix2pix & $38.383^{\circ}$ & 0.592 & 0.469 \\
    sketch2normal & $25.794^{\circ}$ & $0.473$& 0.343 \\
    Ours &  $\mathbf{23.639}^{\circ}$ & $\mathbf{0.443}$ & $\mathbf{0.327}$ \\
    
    \end{tabular}
    \end{center}
    \end{table}

\section{Conclusion}

In this paper, we proposed a novel framework for generating accurate and high-quality normal maps from freehand sketch images. 
The proposed network can generate normal maps with clear boundaries and smooth normal textures. We compared the generated results with three metrics: L1 and L2 distances, and angular difference in a quantitative way. To verify the functions of the proposed geometric sampling, we compare the generated results with different amounts of selected point hints by adjusting the threshold value of curvature to generate high qualified normal map. We set various values of threshold and trained our model to generate normal maps with the same metrics. Finally, the effective mask is adopted to generate the proper distribution of normal map. The generated results show that the proposed method outperforms the previous solutions for normal map generation.

Our proposed approach has some limitations to be improved. First, our generative network can only cope with sketch images of 256$\times$256 resolution limited by the network depth and architecture. Also, the network may fail to generate and fill sketch regions correctly with complex sketch inputs. Therefore, we would like to implement a coarse-fine network structure to deal with higher resolution generation and generate normal maps for complex sketches. In addition, we think that the idea of multi-scale processing of the input sketch \cite{hudon2018deep} may be helpful to achieve higher quality normal maps generation. In this work, we integrated the U-Net architecture into the discriminator for more accurate results, this approach may be improved to achieve better generation results.

\section*{Acknowledgement} 
This work was supported by Grant from Tateishi  Science and Technology Foundation, JAIST Research Grants for Science Hub, and JSPS KAKENHI grant JP20K19845, Japan.

{\small
\bibliographystyle{ieee_fullname}
\bibliography{egbib}
}

\end{document}